\documentclass[conference]{IEEEtran}
\IEEEoverridecommandlockouts
\usepackage{cite}
\usepackage{dblfloatfix}
\usepackage{amsmath,amssymb,amsfonts}
\usepackage{algorithmic}
\usepackage{hyperref}
\hypersetup{
    colorlinks=true,
    linkcolor=blue,
    filecolor=black,      
    urlcolor=blue,
}
\usepackage{graphicx}
\usepackage{textcomp}
\usepackage{xcolor}
\usepackage{multirow}
\def\BibTeX{{\rm B\kern-.05em{\sc i\kern-.025em b}\kern-.08em
    T\kern-.1667em\lower.7ex\hbox{E}\kern-.125emX}}

\begin{document}

\title{Classification of Cuisines from  Sequentially Structured Recipes}

\author{
\IEEEauthorblockN{{ Tript Sharma}}
\IEEEauthorblockA{\textit{Department of Mechanical Engineering} \\
\textit{Delhi Technological University}\\
Delhi, India \\
triptsharma22@gmail.com}
\and
\IEEEauthorblockN{{ Utkarsh Upadhyay}}
\IEEEauthorblockA{\textit{Department of Electronics Engineering} \\
\textit{Jamia Millia Islamia University}\\
Delhi, India\\
utkarshdhy@gmail.com}
\and
\IEEEauthorblockN{Ganesh Bagler}
\IEEEauthorblockA{\textit{Department of Computational Biology} \\
\textit{Indraprastha Institute of Information}\\
\textit{Technology Delhi (IIIT-Delhi)}\\
bagler@iiitd.ac.in}
}

\maketitle

\begin{abstract}
Cultures across the world are distinguished by the idiosyncratic patterns in their cuisines. These cuisines are characterized in terms of their substructures such as ingredients, cooking processes and utensils. A complex fusion of these substructures intrinsic to a region defines the identity of a cuisine. Accurate classification of cuisines based on their culinary features is an outstanding problem and has hitherto been attempted to solve by accounting for ingredients of a recipe as features. Previous studies have attempted cuisine classification by using unstructured recipes without accounting for details of cooking techniques. In reality, the cooking processes/techniques and their order are highly significant for the recipe's structure and hence for its classification. In this article, we have implemented a range of classification techniques by accounting for this information on the RecipeDB dataset containing sequential data on recipes. The state-of-the-art RoBERTa model presented the highest accuracy of 73.30\% among a range of classification models from Logistic Regression and Naive Bayes to LSTMs and Transformers.

\end{abstract}


\section{Introduction}
Cuisines represent the culinary imprint of cultures. The structure of cuisines is shaped by composition of their recipes. Increasing availability of data of cuisines has led to data-driven explorations of cuisines such as food pairing, culinary fingerprinting and cuisine classification. Classification of cuisines is an interesting problem with applications for recipe  recommendation and generation of novel recipes. Hitherto, cuisine classification has been attempted using ingredients of recipes as a feature. These approaches have overlooked key factors such as cooking techniques and their order which clearly form a key aspect of recipes.


We account for the loss of these information by including these details for the cuisine classification problem. As opposed to treating recipes as an itemset, we propose a methodology that views the problem as a Text Classification (TC) problem. TC refers to annotating text with one category or other based on content words and their collocations. In this article, we propose different architectures for cuisine classification on sequential datasets.


\section{Related Work}
\label{relwork}
Availability of detailed recipe data has evoked the interest in recipe recommendation and generation. In the past, classification of cuisines has been attempted on the basis of various factors such as time, ethnicity and place by creating six different feature stylistic sets from the data document~\cite{0}. Han Su et. al.~\cite{1} have worked on cuisine identification by using ingredients used in recipes as a basis. 
By identifying ingredients as features, they could provide insights on cuisine similarity. A personalised cuisine recommendation system based on user's preferences has also been proposed~\cite{2} where user's preferences are derived from their browsing activities. 

Support Vector Machines~\cite{svm4} and several other machine learning techniques have also been implemented towards generation of a cuisine. Recently, a study on classification of cuisine on the basis of the recipe's ingredients~\cite{5} suggested a detailed relation between a recipe and its ingredients. 

In this article, we propose that, beyond ingredients, even the processes and utensils involved in cooking a recipe and their order of occurrence can provide significant insights into the cuisine. We have used RecipeDB ~(site)\cite{recipedb} dataset. And to test our hypothesis, we perform classification on the dataset using several machine learning techniques, neural networks and transformers.

\def\arraystretch{1.3}
\begin{table*}[!t]
\caption{Sample Dataset from RecipDB}
\begin{tabular}{|l|l|l|l|}
\hline
\multicolumn{1}{|c|}{Recipe ID} &
  \multicolumn{1}{c|}{Continent} &
  \multicolumn{1}{c|}{Cuisine} &
  \multicolumn{1}{c|}{Recipe} \\ \hline
2610 &
  African &
  Middle Eastern &
  {[}'water', 'red lentil', 'rom tomato', …,'smooth', 'stir', 'heat'{]} \\ \hline
3957   & Asian          & Southeast Asian     & {[}'olive oil', 'onion', 'garlic', 'ginger', …, 'stir', 'add', 'cook ', 'season', 'garnish', 'pot'{]} \\ \hline
4153   & Asian          & Indian Subcontinent & {[}'coconut milk', 'milk', 'white sugar', 'basmati rice', …, 'stir', 'cook', 'saucepan', 'bowl'{]}    \\ \hline
79897 &
  Latin American &
  Mexican &
  {[}'beef', 'chunky salsa', 'mushroom', 'garlic', …, 'heat', 'simmer', 'serve', 'skillet'{]} \\ \hline
138976 &
  European &
  Deutschland &
  {[}'oven buttermilk biscuit', 'onion', 'cream', ..., 'spread', 'sprinkle', 'bake', 'pan'{]} \\ \hline
149191 & North American & Canadian            & {[}'raisin', 'fig', 'water', 'date', 'butter', …, 'chill', 'cut', 'bowl', 'processor', 'pan'{]}       \\ \hline
\end{tabular}
\label{tab:sample}
\end{table*}

\section{Dataset}
\label{dataset}
\href{https://cosylab.iiitd.edu.in/recipedb}{RecipeDB} was used as the source of structured data on recipes for the analysis. The dataset contains 118,071 recipes obtained from sources like \href{https://www.allrecipes.com}{AllRecipes}, \href{https://www.epicurious.com}{Epicurious}
\href{https://www.foodnetwork.com}{Food Network}, and \href{https://www.tarladalal.com} {TarlaDalal}. The dataset consists of 26 cuisines as shown in Table \ref{tab:db}. Moreover, it contains an aggregation of 20280 unique ingredients, 256 unique processes  and 69 unique utensils. Sample dataset of RecipeDB can be seen in Table \ref{tab:sample}. Our analysis involves the following substructures of cooking recipes pertaining to traditional recipes, namely, recipes, ingredients, processes and utensils.

\def\arraystretch{1.3}
\begin{table}
\centering
\caption{Dataset Information}
\begin{tabular}{|l|l|l|l|} 
\hline
Cuisine               & \begin{tabular}[c]{@{}l@{}}Number of \\Recipes\end{tabular} & Cuisine                                                         & \begin{tabular}[c]{@{}l@{}}Number of \\Recipes\end{tabular}  \\ 
\hline
Australian            & 5823                                                        & Japanese                                                        & 2041                                                         \\ 
\hline
Belgian               & 1060                                                        & Korean                                                          & 668                                                          \\ 
\hline
Canadian              & 6700                                                        & Mexican                                                         & 14463                                                        \\ 
\hline
Caribbean             & 3026                                                        & Middle Eastern                                                  & 3905                                                         \\ 
\hline
Central American      & 460                                                         & Northern Africa                                                 & 1611                                                         \\ 
\hline
Chinese and Mongolian & 5896                                                        & Rest Africa                                                     & 2740                                                         \\ 
\hline
Deutschland           & 4323                                                        & Scandinavian                                                    & 2811                                                         \\ 
\hline
Eastern European      & 2503                                                        & South American                                                  & 7176                                                         \\ 
\hline
French                & 6381                                                        & Southeast Asian                                                 & 1940                                                         \\ 
\hline
Greek                 & 4185                                                        & \begin{tabular}[c]{@{}l@{}}Spanish and\\Portuguese\end{tabular} & 2844                                                         \\ 
\hline
Indian Subcontinent   & 6464                                                        & Thai                                                            & 2605                                                         \\ 
\hline
Irish                 & 2532                                                        & UK                                                              & 4401                                                         \\ 
\hline
Italian               & 16582                                                       & US                                                              & 5031                                                         \\
\hline
\end{tabular}
\label{tab:db}
\end{table}

\def\arraystretch{1.3}
\begin{table}[]
\caption{Frequency Distribution of Features}
\begin{tabular}{|l|l|l|l|}
\hline
Number of Features & Frequency          & Number of Features & Frequency \\ \hline
304                & \textgreater 1000  & \textless 2        & 11738     \\ \hline
106                & \textgreater 5000  & \textless 3        & 14015     \\ \hline
57                 & \textgreater 10000 & \textless 4        & 15002     \\ \hline
43                 & \textgreater 15000 & \textless 5        & 15620     \\ \hline
34                 & \textgreater 20000 & \textless 6        & 16073     \\ \hline
24                 & \textgreater 25000 & \textless 7        & 16394     \\ \hline
19                 & \textgreater 30000 & \textless 8        & 16627     \\ \hline
17                 & \textgreater 35000 & \textless 10       & 17016     \\ \hline
13                 & \textgreater 40000 & \textless 15       & 17314     \\ \hline
12                 & \textgreater 45000 & \textless 20       & 17519     \\ \hline
\end{tabular}
\label{tab:freq}
\end{table}

RecipeDB consists of ingredients, processes and utensils mined from unstructured recipe scraped from the above mentioned resources. The substructures for the recipes are mined in a sequential fashion depending upon the order in which they are used in preparing the dish. 
The dataset is highly sparse with a sparsity ratio of 99.50\%. Out of the 20,400 distinct ingredients obtained, 11738  occur at most in one recipe such as `lasagna noodle wheat', while `add' appeared 1,88,004 number of times. The corresponding cumulative frequency table for the number of items shown in Table~\ref{tab:freq}, represents the nature of the dataset. 

\section{Preprocessing}
\label{preprocess}
Before performing cuisine classification on RecipeDB data, preprocessing was implemented on structured and sequential lists of ingredients, processes and utensils. 
Furthermore, the digits or symbols were omitted from the items to only keep words, thereby reducing the noise in this highly sparse dataset. The preprocessing further involved tokenization followed by lemmatization of the dataset, resulting in 20,400 distinct entities.

The data is further processed to conform to the classification model requirements. Since an individual word itself doesn't impart any semantic or syntactic significance to the classification models that require quantified features as inputs, each item was translated to vectors using two techniques, namely, TF-IDF vectorization and word embedding. Depending upon the preprocessing method used, the models employed can be broadly classified into two categories: sequential models and statistical models. If the dataset is sequential, it is evident that sequential models like RNNs work better while for non sequential datasets, models like Logistic Regression, SVM, etc. perform better.

Word embeddings are essentially word representation as vectors such that semantically similar words have similar vectors whereas TF-IDF vectorization method observes the sequence of items as distinct words. Thus, TF-IDF vectors don't preserve the sequential nature of the data. Yet, we used TF-IDF technique because of its weighted function which reduces the effect of high frequency yet less meaningful words and provides a good analytical cause.



\section{Classification}
\label{classifier}
Classifying the recipes region/cuisine-wise, based on the elements involved in cooking a recipe is a major and the most important part of our analysis. The analysis treats recipes either as a sequential or as an unordered set of items. Many state-of-the-art machine learning models with TF-IDF vector inputs, such as Naive Bayes, Logistic Regression, Ensemble models along with boosting and Support Vector Machines, were tested for text classifications. Also, sequence models such as Recurrent Neural Networks and state-of-the-art NLP transformers such as BERT and RoBERTa were tested on our dataset to analyse the `sequential nature' of the dataset. We will further  discuss the implementations of these classifiers in the `Experiments' section.

\subsection{Naive Bayes}
Naive Bayes (NB) classifier is probabilistic in nature. It is based on the supposition that all the features are independent and autonomous. NB selects the label which maximizes the posterior probability:

\begin{equation}
    P(C_k|x) = P(C_k)*P(x|C_k) / P(x)
\end{equation} 

 while the naive supposition is:
\begin{equation}
    P(x_i|x_{i+1},…,x_n,C_k) = P(X_i|C_k)
\end{equation}

In spite of the fact that the naive supposition is false most of the time, NB gave extremely competitive results with respect to other classifiers.

\subsection{Logistic Regression}

Logistic Regression (LG) is the most used and a fundamental classifier. LG is also probabilistic in nature. It is based on the following Sigmoidal equation:

\begin{equation}
    S(x) = 1/(1 + e^{-x})
\end{equation}

LG treats the problem as a generalized linear regression model, which can be expressed as: 

\begin{equation}
    f(k,i)=\beta_{0,k}+\beta_{1,k}x_{1,i}+ \beta_{2,k}x_{2,i}+\cdots +\beta_{M,k}x_{M,i},
\end{equation}

Here, for our multi-class classification problem LG is trained on a one-vs-rest scheme. Similar steps were followed in a previous research on a different dataset~\cite{5} where LG presented with the best results. This hold true in our cse as well in comparison with other baseline models.

\subsection{Support Vector Machine}
Support Vector Machines (SVMs) have been demonstrated to be having the best performance when working with textual data~\cite{svm4}. For classification, SVMs require translation of a multi-class classification problem to binary classification. For this, the One-vs-All approach was used. Single classifier per class was trained with the training set belonging to that class annotated as positive while the rest of the samples as negative. A strong real-valued confidence score along with a class label, by the base classifiers is required for the decision. SVM then searches for the two best-fit parallel hyperplanes which separates the two classes of data, so that they are farthest from each other. 





\subsection{Random Forest with Boosting}
 Random forest (RF) is a bagging decision tree approach~\cite{rf}. When used as a classifier, it might not perform that well when working with a small number of features. But given that our problem is characterized with a large number of features, techniques such as RF with AdaBoost can turn out to be a good text classifier.

\begin{table*}[htbp]

\caption{Performance Metrics of Applied Models}
\begin{center}

\begin{tabular}{|l|l|l|l|l|l|l|l|l|}
\hline
\multicolumn{1}{|c|}{\multirow{2}{*}{Dataset}} & \multicolumn{1}{c|}{\multirow{2}{*}{\begin{tabular}[c]{@{}c@{}}Performance\\ Metric\end{tabular}}} & \multicolumn{1}{c|}{\multirow{2}{*}{LogReg}} & \multicolumn{1}{c|}{\multirow{2}{*}{Naive Bayes}} & \multicolumn{1}{c|}{\multirow{2}{*}{\begin{tabular}[c]{@{}c@{}}SVM\\ (linear)\end{tabular}}} & \multicolumn{1}{c|}{\multirow{2}{*}{Random Forest}} & \multicolumn{1}{c|}{\multirow{2}{*}{LSTM}} & \multicolumn{2}{c|}{Transformer}                         \\ \cline{8-9} 
\multicolumn{1}{|c|}{}                         & \multicolumn{1}{c|}{}                                                                              & \multicolumn{1}{c|}{}                        & \multicolumn{1}{c|}{}                             & \multicolumn{1}{c|}{}                                                                        & \multicolumn{1}{c|}{}                               & \multicolumn{1}{c|}{}                      & \multicolumn{1}{c|}{BERT} & \multicolumn{1}{c|}{RoBERTa} \\ \hline
\multirow{5}{*}{RecipeDB}                      & Accuracy                                                                                           & 57.70                                        & 51.64                                             & 56.60                                                                                        & 50.37                                               & 53.61                                      & 68.71                     & 73.30                        \\ \cline{2-9} 
                                               & Loss                                                                                               & 1.51                                         & 7.14                                              & 2.97                                                                                         & 2.32                                                & 1.65                                       & 0.21                      & 0.10                         \\ \cline{2-9} 
                                               & Precision                                                                                          & 0.56                                         & 0.50                                              & 0.54                                                                                         & 0.48                                                & 0.53                                       & 0.58                      & 0.67                         \\ \cline{2-9} 
                                               & Recall                                                                                             & 0.57                                         & 0.51                                              & 0.56                                                                                         & 0.50                                                & 0.54                                       & 0.60                      & 0.71                         \\ \cline{2-9} 
                                              & F1 Score                                                                                           & 0.56                                         & 0.50                                              & 0.54                                                                                         & 0.49                                                                                          & 0.53                  & 0.57                      & 0.69
                        \\ \hline
\end{tabular}
\end{center}
\label{tab:models}
\end{table*}

\subsection{RNN (LSTM)}
Recurrent Neural Networks follow a temporal or sequential connection between nodes of a layer~\cite{rnn}. Therefore, they are an upgrade to the conventional neural networks which consider mutual independence among the sequential inputs. Furthermore, RNNs contain an internal `memory' and hence making them suitable for remembering previous inputs. Therefore, the characteristics of RNNs align with our problem. 

We employed a state-of-the-art RNN, the Long-Short Term Memory based neural network (LSTM)~\cite{sundermeyer2012lstm}. LSTMs are more complex than simple RNNs as they involve a cell- and gate-like input, output and forget gate. Using these gates it controls the flow of information through the temporal dimension. It decides whether any piece of information is significant for the broader or immediate goal, or should it be removed from the memory. On account of this significant characteristic, we employed a simple 2-layer LSTM.


\subsection{Transformer}
RNNs go through words in a temporal fashion and if the sequence is long as in case of RecipeDB, the model tends to forget the crucial features of sequentially distant features. In order to overcome this limitation, attention based transformers were developed. Transformers~\cite{vaswani2017attention} are the NLP models which are used to boost the speed of attention based models by enabling parallelization. They completely eliminate the recurrence with self attention to establish relationship between input and output, thus making them suitable for multiple language processing applications.

We employed BERT-base~\cite{bert} and RoBERTa~\cite{roberta} models on RecipeDB. Both models perform bidirectional encoding implementing transformers after pre-training with the exception that RoBERTa is trained differently. RoBERTa was trained on longer sequences for more training steps than BERT.

\section{Experiments}
For the purpose of cuisine classification we implemented different machine learning models. 
We tested the accuracy of the models on the RecipeDB dataset to validate the results obtained. The data was divided into 7:1:2 ratio to obtain training, validation and testing datasets respectively. Therefore out of 1,18,071 recipes training, validation and testing datasets consist of 82,650, 12,021 and 23,380 recipes respectively.

Since recipes were represented as sequences of ingredients, processes and utensils all concatenated together, long sequence were generated. The sequences were pre-processed differently for statistical models and sequential models as described in Section~\ref{preprocess}. By feeding the features obtained as input to the classifiers mentioned in Section~\ref{classifier} yielded results shown in Table~\ref{tab:models}. The corresponding code and relevant files are present in the \href{https://github.com/cosylabiiit/cuisine-classification}{GitHub repository: https://github.com/cosylabiiit/cuisine-classification}.


Among the various statistical models that were implemented for cuisine classification, Logistic Regression performed the best, but with an accuracy of only 57.70\%. These models learn the frequency of occurrence of an ingredient or process or utensil to obtain the features unique to a cuisine instead of treating the recipes as an interrelationship among these items. Furthermore, since the dataset is sparse the models couldn't fit better, leading to high bias. 

Owing to the temporal relationship among the items in recipes, we observed that sequential models perform better than the statistical models. However, LSTM model gave a lower accuracy than Logistic Regression and Linear kernel SVM used in~\cite{5}. The lower accuracy is justified as the model is among the most simplistic models in the recurrent neural network class. Furthermore while comparing the LSTM model with Transformers, the sequences are treated differently i.e. LSTMs consider left to right sequence order unlike the bi-directional check in Transformers. Moreover, despite having better memory logic than vanilla RNNs, LSTMs are limited by the number of words in the sequence which further reduces the accuracy. 

The limitations of LSTMs have been overcome in Transformers as explained earlier which resulted in the optimal accuracy of 73.3\% and a loss of 0.10 on the RecipeDB dataset. Hence, the model is able to predict the class with least errors on minimum number of datasets among the models tested. 
The model presents a high average precision, recall and F1 score values representing its  ability for cuisine classification.

\section{Conclusions}
This article investigates different approaches for cuisine classification as a synthesis of ingredients, processes and utensils inherent to a cuisine. It also examines the effect of temporal relationships among the features to fingerprint the worldwide cuisines with the state-of-the-art RoBERTa model giving optimal results for the problem. Thus, we present a strategy to treat recipes as chains of events that are similar for a region and simultaneously contrasting from others to some extent to enable classification. Further, this articles has raised issues that can help optimise the results in different computational contexts such as recipe generation and recipe recommendation.

Apart from this, the article also raises some new research questions relating to cuisine classification. While our analysis considered for the sequential nature of recipes, the relationship among the three substructures remains unaccounted.  Moreover, what features aid or hinder the classification of a recipe which could help one to uniquely distinguish between the cuisines? While maintaining the sequential nature of the recipes, redundant features were not removed. Hence, future analysis needs to identify the effect induced by these features on the classification accuracy of the models. Furthermore the imbalance among the classes affects the cuisine prediction accuracy of the classifiers. This can be reduced by ignoring the low frequency classes but would lead to a limited exploration of the world cuisines. This  trade-off presents as a  dilemma in this analysis.

We believe this article adds another dimension to the existing body of research on cuisine classification. This is of value for cuisine classification of unknown recipes and also aids in identifying salient features intrinsic to a cuisine. 

\section{Acknowledgement}
G.B. thanks the Indraprastha Institute of Information
Technology (IIIT-Delhi) for providing computational facilities and support. T.S and U.U. are Research Interns in Dr. Bagler's lab (Complex Systems Laboratory) at the Center for Computational Biology and thank  IIIT-Delhi for the support.

\bibliographystyle{biblo_name}

\bibliography{biblo}

\end{document}